\documentclass[10pt,twocolumn,english]{article}
\renewcommand{\rmdefault}{cmr}

\usepackage[T1]{fontenc}
\usepackage[latin9]{inputenc}
\usepackage{xcolor}
\usepackage{url}
\usepackage{amsmath}
\usepackage{amssymb}
\usepackage{graphicx}

\makeatletter

\providecommand{\tabularnewline}{\\}

\newenvironment{lyxlist}[1]
	{\begin{list}{}
		{\settowidth{\labelwidth}{#1}
		 \setlength{\leftmargin}{\labelwidth}
		 \addtolength{\leftmargin}{\labelsep}
		 }}
	{\end{list}}

\usepackage{spconf, graphicx}
\usepackage{pgfplots}
\usepackage{tikz}

\name{S. Pi\'erard, and M. Van Droogenbroeck}
\address{Department of Electrical Engineering and Computer Science, University of Li\`ege, Belgium \\ \{S.Pierard, M.VanDroogenbroeck\}@uliege.be}

\usepackage{xspace}

\newcommand{\etc}{etc\xspace}
\newcommand{\changedetection}{CDNET\xspace}

\newtheorem{randomExperiment}{Random Experiment}

\@ifundefined{showcaptionsetup}{}{%
 \PassOptionsToPackage{caption=false}{subfig}}
\usepackage{subfig}
\makeatother

\usepackage{babel}
\begin{document}
\global\long\def\tn{\mathrm{tn}}%

\global\long\def\fp{\mathrm{fp}}%

\global\long\def\fn{\mathrm{fn}}%

\global\long\def\tp{\mathrm{tp}}%

\global\long\def\numTN{\mathsf{TN}}%

\global\long\def\numFP{\mathsf{FP}}%

\global\long\def\numFN{\mathsf{FN}}%

\global\long\def\numTP{\mathsf{TP}}%

\global\long\def\proba#1{P\left(#1\right)}%

\global\long\def\clazz{y}%

\global\long\def\cpos{c^{+}}%

\global\long\def\cneg{c^{-}}%

\global\long\def\estimate#1{\widehat{#1}}%

\global\long\def\metric{\text{I}}%

\global\long\def\ppv{\text{PPV}}%

\global\long\def\npv{\text{NPV}}%

\global\long\def\tpr{\text{TPR}}%

\global\long\def\tnr{\text{TNR}}%

\global\long\def\fprE{\mbox{\ensuremath{\left(1-\tnr\right)}}}%

\global\long\def\fpr{\text{FPR}}%

\global\long\def\fnr{\mbox{\ensuremath{\left(1-\tpr\right)}}}%

\global\long\def\fprx{\text{FPR}}%

\global\long\def\fnrx{\text{FNR}}%

\global\long\def\accuracy{\text{A}}%

\global\long\def\errorrate{\text{ER}}%

\global\long\def\balancedaccuracy{\text{BA}}%

\global\long\def\jaccard{\text{J}}%

\global\long\def\auc{\text{AUC}}%

\global\long\def\bias{\text{B}}%

\global\long\def\priorSymbol{\pi}%

\global\long\def\priorneg{\priorSymbol{}^{-}}%

\global\long\def\priorpos{\priorSymbol^{+}}%

\global\long\def\rateOfNegativePredictions{\tau^{-}}%

\global\long\def\rateOfPositivePredictions{\tau^{+}}%

\global\long\def\setofcontexts{\Gamma}%

\global\long\def\context{\gamma}%

\global\long\def\weight{\omega}%

\global\long\def\firstSet{\mathcal{A}}%

\global\long\def\secondSet{\mathcal{B}}%

\global\long\def\comma{\enspace\mbox{,}}%

\global\long\def\dot{\enspace\mbox{.}}%

\global\long\def\Fscore{\text{F}}%

\global\long\def\recall{\text{R}}%

\global\long\def\precision{\text{P}}%

\global\long\def\cardinality{\#}%


\global\long\def\prediction{\hat{y}}%

\global\long\def\randomVariablePrediction{\hat{Y}}%

\global\long\def\groundtruth{y}%

\global\long\def\randomVariableGroundTruth{Y}%

\global\long\def\comparison{\delta}%

\global\long\def\randomVariableComparison{\Delta}%


\global\long\def\indicator{\text{I}}%

\global\long\def\summarization#1{#1}%

\global\long\def\arithaverage#1{#1}%

\global\long\def\arithmeticMean#1{\bar{#1}}%

\global\long\def\cardinalityOf#1{\#_{#1}}%


\global\long\def\selectionProbability#1{\proba{#1}}%

\global\long\def\selectionIndex{v}%

\global\long\def\selectionRandomVariable{V}%

\global\long\def\setOfSources{\mathbb{V}}%

\global\long\def\numberOfSources{N}%

\title{Summarizing the performances of a background subtraction algorithm
measured on several videos}
\maketitle
\begin{abstract}
There exist many background subtraction algorithms to detect motion
in videos. To help comparing them, datasets with ground-truth data
such as \changedetection or LASIESTA have been proposed. These datasets
organize videos in categories that represent typical challenges for
background subtraction. The evaluation procedure promoted by their
authors consists in measuring performance indicators for each video
separately and to average them hierarchically, within a category first,
then between categories, a procedure which we name ``summarization''.
While the summarization by averaging performance indicators is a valuable
effort to standardize the evaluation procedure, it has no theoretical
justification and it breaks the intrinsic relationships between summarized
indicators. This leads to interpretation inconsistencies. In this
paper, we present a theoretical approach to summarize the performances
for multiple videos that preserves the relationships between performance
indicators. In addition, we give formulas and an algorithm to calculate
summarized performances. Finally, we showcase our observations on
\changedetection 2014.

\end{abstract}

\section{Introduction}

A plethora of background subtraction algorithms have been proposed
in the literature~\cite{Bouwmans2010Statistical,Bouwmans2014Traditional,Bouwmans2019Deep,Elhabian2008Moving,Jodoin2014Overview,Sobral2014AComprehensive,Vaswani2018Robust}.
They aim at predicting, for each pixel of each frame of a video, whether
the pixel belongs to the background, free of moving objects, or to
the foreground. Background subtraction algorithms have to operate
in various conditions (viewpoint, shadows, lighting conditions, camera
jitter, \etc). These conditions are covered by different videos in
evaluation datasets, such as \changedetection~\cite{Goyette2012Changedetection,Goyette2014ANovel}
or LASIESTA~\cite{Cuevas2016Labeled}.

Measuring the performance on a single video is done at the pixel level,
each pixel being associated with a ground-truth class $\groundtruth$
(either negative $\cneg$ for the background, or positive $\cpos$
for the foreground) and an estimated class $\prediction$ (the temporal
and spatial dependences between pixels are ignored). Performance indicators
adopted in the field of background subtraction are mainly those used
for two-class crisp classifiers (precision, recall, sensitivity, F-score,
error rate, \etc).

To compare the behavior of algorithms, it is helpful to ``\emph{summarize}''
all the performance indicators that are originally measured on the
individual videos. However, to the best of our knowledge, there is
no theory for this summarization process, although an attempt to standardize
it has been promoted with \changedetection. In \changedetection,
videos are grouped into 11 categories, all videos in a given category
having the same importance, and all categories having also the same
importance. In other words, the videos are weighted regardless of
their size. The standardized summarization process of \changedetection
is performed hierarchically by computing an arithmetic mean, first
between the videos in a category, then between the categories. This
averaging is done for seven performance indicators, one by one, independently.
While the procedure of \changedetection is valuable, it has two major
drawbacks related to the interpretability of the summarized values.
\begin{figure}
\begin{centering}
\resizebox{1\columnwidth}{!}{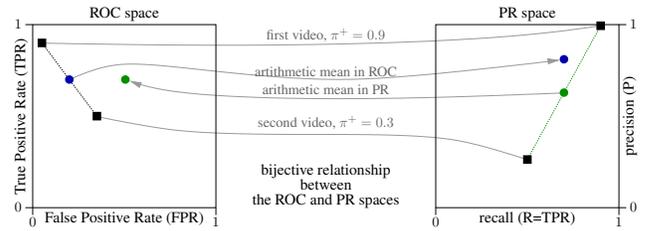}
\par\end{centering}
\caption{How can we summarize the performances of a background subtraction
algorithm, measured on several videos? This figure shows that arithmetically
averaging the performances breaks the bijective relationship that
exists between the ROC and Precision-Recall ($\protect\precision\protect\recall$)
spaces~\cite{Davis2006TheRelationship}. This paper presents a new
summarization technique that guarantees the coherence between any
sets of performance indicators.\label{fig:graphical-abstract}}
\end{figure}

First, the procedure breaks the intrinsic relationships between performance
indicators. The case of the M4CD algorithm, as evaluated on \changedetection,
topically illustrates this inconsistency. Despite that the F-score
is known to be the harmonic mean of precision and recall, the arithmetic
mean of the F-scores across all videos is $0.69$ while the harmonic
mean between the arithmetically averaged precisions and recalls is
$0.75$. The difficulty to summarize F-scores has already been discussed
in~\cite{Forman2010Apples}, when averaging between folds in cross-validation.
Such a problem also occurs for other indicators. Moreover, Figure~\ref{fig:graphical-abstract}
shows that the summarization with the arithmetic mean breaks the bijective
relationship between the ROC and $\precision\recall$ spaces~\cite{Davis2006TheRelationship}.
It is not equivalent to summarize the performances in one space or
the other. In particular, one could obtain a meaningless arithmetically
averaged performance point, located in the unachievable part of the
PR space~\cite{Boyd2012Unachievable} (the achievable part is not
convex). This leads to difficulties for the interpretation and, eventually,
to contradictions between published summarized results.

Second, some indicators (such as the precision, recall, sensitivity,
or error rate) have a probabilistic meaning~\cite{Goutte2005AProbabilistic}.
Arithmetically averaging these indicators does not lead to a value
that preserves the probabilistic meaning. This leads to interpretability
issues. Strictly speaking, one cannot think in terms of probabilities
unless a random experiment is defined very precisely, as shown by
Bertrand's paradox~\cite{Bertrand1889Calcul}.

This paper presents a better summarization procedure that avoids these
interpretability issues. In Section~\ref{sec:one-video-sequence},
we present indicators to measure the performance of an algorithm applied
on a single video and pose the random experiment that underpins the
probabilistic meaning of these indicators. Then, in Section~\ref{sec:several-video-sequences},
we generalize this random experiment for several videos, and show
that the resulting performance indicators can be computed based only
on the indicators obtained separately for each video. In other words,
we have established a theoretical model for summarization that guarantees
a consistency between the indicators resulting from both random experiments.
In Section~\ref{sec:application-to-bgs}, we showcase our new summarization
procedure on \changedetection and discuss how it affects the ranking
between algorithms. Section~\ref{sec:conclusion} concludes the paper.

\section{Performance indicators for one video}

\label{sec:one-video-sequence}

The performance of a background subtraction algorithm on a video is
measured by running it once (even for non-deterministic algorithms),
and counting at the pixel level the amounts of true negatives $\numTN$
($\groundtruth=\prediction=\cneg$), false positives $\numFP$ ($\groundtruth=\cneg$
and $\prediction=\cpos$), false negatives $\numFN$ ($\groundtruth=\cpos$
and $\prediction=\cneg$), and true positives $\numTP$ ($\groundtruth=\prediction=\cpos$).

The performance indicators are then derived from these amounts. For
example, the prior of the positive class is $\priorpos=\frac{\numFN+\numTP}{\numTN+\numFP+\numFN+\numTP}$,
the rate of positive predictions is $\rateOfPositivePredictions=\frac{\numFP+\numTP}{\numTN+\numFP+\numFN+\numTP}$,
the error rate is $\errorrate=\frac{\numFP+\numFN}{\numTN+\numFP+\numFN+\numTP}$,
the true negative rate (specificity) is $\tnr=\frac{\numTN}{\numTN+\numFP}$,
the false positive rate is $\fpr=\frac{\numFP}{\numTN+\numFP}$, the
false negative rate is $\fnrx=\frac{\numFN}{\numFN+\numTP}$, the
true positive rate (recall) is $\tpr=\recall=\frac{\numTP}{\numFN+\numTP}$,
the positive predictive value (precision) is $\ppv=\precision=\frac{\numTP}{\numFP+\numTP}$,
and the F-score is $\Fscore=\frac{2\numTP}{\numFP+\numFN+2\numTP}$.

The $(\fprx,\,\tpr)$ coordinates define the well-known Receiver Operating
Characteristic (ROC) space~\cite{Flach2003TheGeometry,Fawcett2006AnIntroduction}.
The precision $\precision$ and recall $\recall$ define the $\precision\recall$
evaluation space~\cite{Boyd2012Unachievable}, which is an alternative
to the ROC space, as there exists a bijection between these spaces,
for a given prior $\priorpos$~\cite{Davis2006TheRelationship}.
In fact, this bijection is just a particular case of the relationships
that exist between the various indicators. There are other famous
relationships. For example, the F-score is known to be the harmonic
mean of the precision $\precision$ and recall $\recall$.

Despite its importance for the interpretation, it is often overlooked
that some indicators have a probabilistic meaning~\cite{Goutte2005AProbabilistic}.
Let us consider the following random experiment.\begin{randomExperiment}[for one video]Draw
one pixel at random (all pixels being equally likely) from the video
and observe the ground-truth class $\randomVariableGroundTruth$ as
well as the predicted class $\randomVariablePrediction$ for this
pixel. The result of the random experiment is the pair $\randomVariableComparison=(\randomVariableGroundTruth,\randomVariablePrediction)$.\end{randomExperiment}We
use capital letters for $\randomVariableGroundTruth$, $\randomVariablePrediction$,
and $\randomVariableComparison$ to emphasize the random nature of
these variables. The outcome of this random experiment can be associated
with a true negative $\tn=(\cneg,\cneg)$, a false positive $\fp=(\cneg,\cpos)$,
a false negative $\fn=(\cpos,\cneg)$, or a true positive $\tp=(\cpos,\cpos)$.
The family of \emph{probabilistic indicators} can be defined based
on this random experiment: 
\begin{equation}
\proba{\randomVariableComparison\in\firstSet\vert\randomVariableComparison\in\secondSet}\textrm{ with }\emptyset\subsetneq\firstSet\subsetneq\secondSet\subseteq\left\{ \tn,\fp,\fn,\tp\right\} 
\end{equation}
It includes $\priorpos=\proba{\randomVariableComparison\in\left\{ \fn,\tp\right\} }$,
$\rateOfPositivePredictions=\proba{\randomVariableComparison\in\left\{ \fp,\tp\right\} }$,
$\errorrate=\proba{\randomVariableComparison\in\left\{ \fp,\fn\right\} }$,
$\tnr=\proba{\randomVariableComparison=\tn\vert\randomVariableComparison\in\left\{ \tn,\fp\right\} }$,
$\tpr=\proba{\randomVariableComparison=\tp\vert\randomVariableComparison\in\left\{ \fn,\tp\right\} }$,
and $\ppv=\proba{\randomVariableComparison=\tp\vert\randomVariableComparison\in\left\{ \fp,\tp\right\} }$.
All the other performance indicators (the F-score, balanced accuracy,
\etc) can be derived from the probabilistic indicators.

\section{Summarizing the performance indicators for several videos}

\label{sec:several-video-sequences}

Let us denote a set of videos by $\setOfSources$. We generalize the
random experiment defined for one video to several videos as follows.
\begin{randomExperiment}[for several videos]First, draw one video
$\selectionRandomVariable$ at random in the set $\setOfSources$,
following an arbitrarily chosen distribution $\proba{\selectionRandomVariable}$.
Then, draw one pixel at random (all pixels being equally likely) from
$\selectionRandomVariable$ and observe the ground-truth class $\randomVariableGroundTruth$
and the predicted class $\randomVariablePrediction$ for this pixel.
The result of the experiment is the pair $\randomVariableComparison=(\randomVariableGroundTruth,\randomVariablePrediction)$.\end{randomExperiment}As
the result of this second random experiment is again a pair of classes,
as encountered for a single video (first random experiment), all performance
indicators can be defined in the same way. This is important for the
interpretability, since it guarantees that all indicators are coherent
and that the probabilistic indicators preserve a probabilistic meaning.

The distribution of probabilities $\proba{\selectionRandomVariable}$
parameterizes the random experiment. It can be arbitrarily chosen
to reflect the relative weights given to the videos. For example,
one could use the weights considered in \changedetection. An alternative
consists in choosing weights proportional to the size of the videos
(the number of pixels multiplied by the number of frames). This leads
to identical distributions $\proba{\randomVariableComparison}$ between
the second random experiment and the first one with the fictitiously
aggregated video $\bigcup_{\selectionIndex\in\setOfSources}\selectionIndex$.

\paragraph*{Summarization formulas.}

\noindent We denote by $\indicator(\selectionIndex)$ the value of
an indicator $\indicator$ obtained with our first random experiment
on a video $\selectionIndex\in\setOfSources$, and by $\indicator(\setOfSources)$
the value of $\indicator$ resulting from our second random experiment
applied on the set $\setOfSources$ of videos.

\noindent We claim that the indicators $\indicator(\setOfSources)$
summarize the performances corresponding to the various videos, because
they can be computed based on some $\indicator(\selectionIndex)$,
exclusively. To prove it, let us consider a probabilistic indicator
$\indicator_{\firstSet\vert\secondSet}$ defined as $\proba{\randomVariableComparison\in\firstSet\vert\randomVariableComparison\in\secondSet}$,
and $\indicator_{\secondSet}$ as $\proba{\randomVariableComparison\in\secondSet}$.
Thanks to our second random experiment, we have
\begin{align}
\indicator_{\firstSet\vert\secondSet}(\setOfSources) & =\proba{\randomVariableComparison\in\firstSet\vert\randomVariableComparison\in\secondSet}\\
 & =\sum_{\selectionIndex\in\setOfSources}\proba{\randomVariableComparison\in\firstSet,\selectionRandomVariable=\selectionIndex\vert\randomVariableComparison\in\secondSet}\\
 & =\sum_{\selectionIndex\in\setOfSources}\proba{\selectionRandomVariable=\selectionIndex\vert\randomVariableComparison\in\secondSet}\,\proba{\randomVariableComparison\in\firstSet\vert\randomVariableComparison\in\secondSet,\selectionRandomVariable=\selectionIndex}\nonumber \\
 & =\sum_{\selectionIndex\in\setOfSources}\proba{\selectionRandomVariable=\selectionIndex\vert\randomVariableComparison\in\secondSet}\,\indicator_{\firstSet\vert\secondSet}(\selectionIndex)\comma
\end{align}
with
\begin{align}
\proba{\selectionRandomVariable=\selectionIndex\vert\randomVariableComparison\in\secondSet} & =\frac{\proba{\selectionRandomVariable=\selectionIndex}\,\proba{\randomVariableComparison\in\secondSet\vert\selectionRandomVariable=\selectionIndex}}{\proba{\randomVariableComparison\in\secondSet}}\\
 & =\frac{\proba{\selectionRandomVariable=\selectionIndex}\,\proba{\randomVariableComparison\in\secondSet\vert\selectionRandomVariable=\selectionIndex}}{\sum_{\selectionIndex'\in\setOfSources}\proba{\selectionRandomVariable=\selectionIndex'}\,\proba{\randomVariableComparison\in\secondSet\vert\selectionRandomVariable=\selectionIndex'}}\nonumber \\
 & =\frac{\proba{\selectionRandomVariable=\selectionIndex}\,\indicator_{\secondSet}(\selectionIndex)}{\sum_{\selectionIndex'\in\setOfSources}\proba{\selectionRandomVariable=\selectionIndex'}\,\indicator_{\secondSet}(\selectionIndex')}\,.
\end{align}
Thus, the summarized probabilistic indicator $\indicator_{\firstSet\vert\secondSet}(\setOfSources)$
is a weighted arithmetic mean of the indicators $\{\indicator_{\firstSet\vert\secondSet}(\selectionIndex)\vert\selectionIndex\in\setOfSources\}$,
the weights being the (normalized) product between the relative importance
$\proba{\selectionRandomVariable=\selectionIndex}$ given to the video
$\selectionIndex$ and the corresponding $\indicator_{\secondSet}(\selectionIndex)$:
\begin{equation}
\indicator_{\firstSet\vert\secondSet}(\setOfSources)=\sum_{\selectionIndex\in\setOfSources}\frac{\proba{\selectionRandomVariable=\selectionIndex}\,\indicator_{\secondSet}(\selectionIndex)}{\sum_{\selectionIndex'\in\setOfSources}\proba{\selectionRandomVariable=\selectionIndex'}\,\indicator_{\secondSet}(\selectionIndex')}\,\indicator_{\firstSet\vert\secondSet}(\selectionIndex).\label{eq:summaization-for-probabilistic-indicators}
\end{equation}
In the particular case of an unconditional probabilistic indicator
$\indicator_{\firstSet}=\indicator_{\firstSet\vert\{\tn,\fp,\fn,\tp\}}$,
Equation~(\ref{eq:summaization-for-probabilistic-indicators}) reduces
to
\begin{equation}
\indicator_{\firstSet}(\setOfSources)=\sum_{\selectionIndex\in\setOfSources}\proba{\selectionRandomVariable=\selectionIndex}\,\indicator_{\firstSet}(\selectionIndex)\,.\label{eq:summarization-formula-for-unconditional-probabilities}
\end{equation}
Note that we are able to summarize probabilistic indicators that are
undefined for some, but not all, videos. The reason is that $\indicator_{\firstSet\vert\secondSet}(\selectionIndex)$
is undefined only when $\indicator_{\secondSet}(\selectionIndex)=0$,
and we observe that only the product of these two quantities appears
in Equation~(\ref{eq:summaization-for-probabilistic-indicators}).
This product is always well defined since it is an unconditional probability
($\indicator_{\secondSet}(\selectionIndex)\,\indicator_{\firstSet\vert\secondSet}(\selectionIndex)=\indicator_{\firstSet\cap\secondSet}(\selectionIndex)$).
To emphasize it, we rewrite Equation~(\ref{eq:summaization-for-probabilistic-indicators})
as
\begin{multline}
\indicator_{\firstSet\vert\secondSet}(\setOfSources)=\sum_{\selectionIndex\in\setOfSources}\frac{\proba{\selectionRandomVariable=\selectionIndex}\,\indicator_{\secondSet}(\selectionIndex)\,\indicator_{\firstSet\vert\secondSet}(\selectionIndex)}{\indicator_{\secondSet}(\setOfSources)}\\
=\frac{\sum_{\selectionIndex\in\setOfSources}\proba{\selectionRandomVariable=\selectionIndex}\,\indicator_{\firstSet\cap\secondSet}(\selectionIndex)}{\indicator_{\secondSet}(\setOfSources)}=\frac{\indicator_{\firstSet\cap\secondSet}(\setOfSources)}{\indicator_{\secondSet}(\setOfSources)}\,.
\end{multline}

\paragraph*{Applying our summarization in the ROC $(\protect\fprx,\protect\tpr)$
and PR $(\protect\recall=\protect\tpr,\protect\precision=\protect\ppv)$
spaces.}

Let us consider the case in which $\priorpos$, $\rateOfPositivePredictions$,
$\fpr$, $\tpr$, and $\ppv$ are known for each video. The summarized
indicators can be computed by Equations~(\ref{eq:summaization-for-probabilistic-indicators})-(\ref{eq:summarization-formula-for-unconditional-probabilities}),
since $\priorpos=\indicator_{\left\{ \fn,\tp\right\} }$, $\rateOfPositivePredictions=\indicator_{\left\{ \fp,\tp\right\} }$,
$\fpr=\indicator_{\left\{ \fp\right\} \vert\left\{ \tn,\fp\right\} }$,
$\tpr=\indicator_{\left\{ \tp\right\} \vert\left\{ \fn,\tp\right\} }$,
and $\ppv=\indicator_{\left\{ \tp\right\} \vert\left\{ \fp,\tp\right\} }$.
This yields:
\begin{align}
\priorpos(\setOfSources) & =\sum_{\selectionIndex\in\setOfSources}\proba{\selectionRandomVariable=\selectionIndex}\,\priorpos(\selectionIndex)\,,\\
\rateOfPositivePredictions(\setOfSources) & =\sum_{\selectionIndex\in\setOfSources}\proba{\selectionRandomVariable=\selectionIndex}\,\rateOfPositivePredictions(\selectionIndex)\,,\\
\fpr(\setOfSources) & =\frac{1}{\priorneg(\setOfSources)}\sum_{\selectionIndex\in\setOfSources}\proba{\selectionRandomVariable=\selectionIndex}\,\priorneg(\selectionIndex)\,\fpr(\selectionIndex)\,,\\
\tpr(\setOfSources) & =\frac{1}{\priorpos(\setOfSources)}\sum_{\selectionIndex\in\setOfSources}\proba{\selectionRandomVariable=\selectionIndex}\,\priorpos(\selectionIndex)\,\tpr(\selectionIndex)\,,\\
\ppv(\setOfSources) & =\frac{1}{\rateOfPositivePredictions(\setOfSources)}\sum_{\selectionIndex\in\setOfSources}\proba{\selectionRandomVariable=\selectionIndex}\,\rateOfPositivePredictions(\selectionIndex)\,\ppv(\selectionIndex)\,.
\end{align}
The passage from ROC to PR is given by:

\begin{equation}
\ppv(\setOfSources)=\frac{\priorpos(\setOfSources)\,\tpr(\setOfSources)}{\priorneg(\setOfSources)\,\fprx(\setOfSources)+\priorpos(\setOfSources)\,\tpr(\setOfSources)}\,.
\end{equation}

\paragraph*{A generic algorithm for the computation of any summarized indicator.}

Let us assume that the elements of the normalized confusion matrix
($\indicator_{\{\tn\}}$, $\indicator_{\{\fp\}}$, $\indicator_{\{\fn\}}$,
and $\indicator_{\{\tp\}}$) can be retrieved for each video. This
could be done by computing them based on other indicators. In this
case, an easy-to-code algorithm to summarize the performances is the
following:
\begin{lyxlist}{00.00.0000}
\item [{step~1:}] arbitrarily weight the videos with $\proba{\selectionRandomVariable}$,
\item [{step~2:}] retrieve $\indicator_{\{\tn\}}(\selectionIndex)$, $\indicator_{\{\fp\}}(\selectionIndex)$,
$\indicator_{\{\fn\}}(\selectionIndex)$, and $\indicator_{\{\tp\}}(\selectionIndex)$
for each video $\selectionIndex$,
\item [{step~3:}] then compute $\indicator_{\{\tn\}}(\setOfSources)$,
$\indicator_{\{\fp\}}(\setOfSources)$, $\indicator_{\{\fn\}}(\setOfSources)$,
and $\indicator_{\{\tp\}}(\setOfSources)$ with Equation~(\ref{eq:summarization-formula-for-unconditional-probabilities}),
\item [{step~4:}] and finally derive all the desired summarized indicators
based on their relationships with the $\indicator_{\{\tn\}},\indicator_{\{\fp\}},\indicator_{\{\fn\}},\indicator_{\{\tp\}}$
indicators. For example, 
\[
\Fscore(\setOfSources)=\frac{2\indicator_{\{\tp\}}(\setOfSources)}{\indicator_{\{\fp\}}(\setOfSources)+\indicator_{\{\fn\}}(\setOfSources)+2\indicator_{\{\tp\}}(\setOfSources)}\,.
\]
\end{lyxlist}

\section{Experiments with \changedetection 2014}

\label{sec:application-to-bgs}

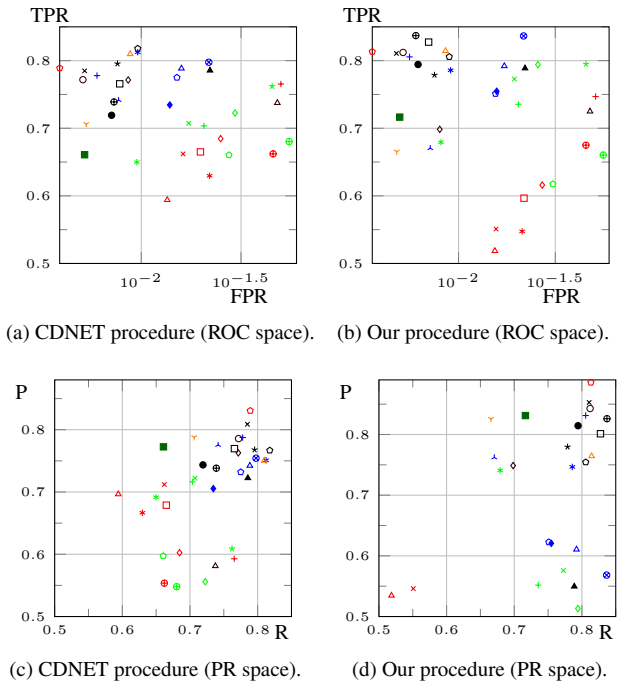
\begin{figure}[!b]
\begin{centering}
\subfloat[\changedetection procedure (ROC space).]{\begin{centering}
\pgfplotsset{
    standard/.style={
        axis x line=middle,
        axis y line=middle,
        enlarge x limits=0.15,
        enlarge y limits=0.15,
        every axis x label/.style={at={(current axis.right of origin)},anchor=north west},
        every axis y label/.style={at={(current axis.above origin)},anchor=north east}
    }
}

\begin{tikzpicture}
\begin{axis}[
xmode=log,
	clip bounding box=upper bound,
	legend pos=south east,
	label style={font=\footnotesize},
	tick style={font=\tiny},
	tick label style={font=\tiny},
	xticklabel style={
		/pgf/number format/.cd,%
          scaled x ticks = false,
          set decimal separator={.},
          fixed},
	xlabel=$\arithaverage{\fpr}$, 
	xlabel style={at={(0.8,0.1)}},
	ylabel=$\arithaverage{\tpr}$, 
	ylabel style={at={(0.35,1.05)},rotate=-90},
	width=0.55\columnwidth,
	height=0.55\columnwidth,
	enlargelimits=false,
	xmin=0, 
	xmax=0.06,
	ymin=0.5, 
	ymax=0.85,
	grid=major,
	minor x tick num=1,
	minor y tick num=1,
	scatter/classes={
		a={mark=x,color=black},
		b={mark=+,color=blue},
		c={mark=o,color=black!80!red},
		d={mark=asterisk,color=blue},
		e={mark=star},
		f={mark=triangle,color=orange},
		g={mark=square},
		h={mark=triangle,color=blue},
		i={mark=diamond,color=black!80!red},
		j={mark=Mercedes star,color=blue},
		k={mark=Mercedes star flipped,color=orange},
		l={mark=pentagon,color=blue},
		m={mark=oplus},
		n={mark=otimes,color=blue},
		o={mark=*},
		p={mark=square*,color=black!60!green},
		q={mark=triangle*},
		r={mark=diamond*,color=blue},
		s={mark=x,color=green},
		t={mark=x,color=red},
		u={mark=+,color=green},
		v={mark=+,color=red},
		w={mark=asterisk,color=green},
		x={mark=asterisk,color=red},
		y={mark=star,color=green},
		z={mark=square,color=red},
		ab={mark=square,color=black!40!green},
		ac={mark=triangle,color=red},
		ad={mark=triangle,color=black!80!red},
		ae={mark=diamond,color=red},
		af={mark=diamond,color=green},
		ag={mark=pentagon,color=red},
		ah={mark=pentagon,color=green},
		ai={mark=oplus,color=red},
		aj={mark=oplus,color=green},
		sbs={mark=pentagon,color=red},
		wise={mark=pentagon}
		}
	]
    \addplot[ 	scatter,only marks,mark options={scale=0.6},
		scatter src=explicit symbolic]
	table[x=FPR,y=R,meta=label] {cdnet-2014.dat};
\end{axis}
\end{tikzpicture}
\par\end{centering}
}\subfloat[Our procedure (ROC space).\label{fig-sub:ROC-ours}]{\begin{centering}
\pgfplotsset{
    standard/.style={
        axis x line=middle,
        axis y line=middle,
        enlarge x limits=0.15,
        enlarge y limits=0.15,
        every axis x label/.style={at={(current axis.right of origin)},anchor=north west},
        every axis y label/.style={at={(current axis.above origin)},anchor=north east}
    }
}

\begin{tikzpicture}
\begin{axis}[
xmode=log,
	clip bounding box=upper bound,
	legend pos=south east,
	label style={font=\footnotesize},
	tick style={font=\tiny},
	tick label style={font=\tiny},
	xticklabel style={
		/pgf/number format/.cd,%
          scaled x ticks = false,
          set decimal separator={.},
          fixed},
	xlabel=$\summarization{\fpr}$, 
	xlabel style={at={(0.8,0.1)}},
	ylabel=$\summarization{\tpr}$, 
	ylabel style={at={(0.35,1.05)},rotate=-90},
	width=0.55\columnwidth,
	height=0.55\columnwidth,
	enlargelimits=false,
	xmin=0, 
	xmax=0.06, 
	ymin=0.5, 
	ymax=0.85,
	grid=major,
	minor x tick num=1,
	minor y tick num=1,
	scatter/classes={
		a={mark=x,color=black},
		b={mark=+,color=blue},
		c={mark=o,color=black!80!red},
		d={mark=asterisk,color=blue},
		e={mark=star},
		f={mark=triangle,color=orange},
		g={mark=square},
		h={mark=triangle,color=blue},
		i={mark=diamond,color=black!80!red},
		j={mark=Mercedes star,color=blue},
		k={mark=Mercedes star flipped,color=orange},
		l={mark=pentagon,color=blue},
		m={mark=oplus},
		n={mark=otimes,color=blue},
		o={mark=*},
		p={mark=square*,color=black!60!green},
		q={mark=triangle*},
		r={mark=diamond*,color=blue},
		s={mark=x,color=green},
		t={mark=x,color=red},
		u={mark=+,color=green},
		v={mark=+,color=red},
		w={mark=asterisk,color=green},
		x={mark=asterisk,color=red},
		y={mark=star,color=green},
		z={mark=square,color=red},
		ab={mark=square,color=black!40!green},
		ac={mark=triangle,color=red},
		ad={mark=triangle,color=black!80!red},
		ae={mark=diamond,color=red},
		af={mark=diamond,color=green},
		ag={mark=pentagon,color=red},
		ah={mark=pentagon,color=green},
		ai={mark=oplus,color=red},
		aj={mark=oplus,color=green},
		sbs={mark=pentagon,color=red},
		wise={mark=pentagon}
		}
	]
    \addplot[ 	scatter,only marks,mark options={scale=0.6},
		scatter src=explicit symbolic]
	table[x=FPRC,y=TPRC,meta=label] {cdnet-2014.dat};
\end{axis}
\end{tikzpicture}
\par\end{centering}
}
\par\end{centering}
\begin{centering}
\subfloat[\changedetection procedure (PR space).]{\begin{centering}
\pgfplotsset{
    standard/.style={
        axis x line=middle,
        axis y line=middle,
        enlarge x limits=0.15,
        enlarge y limits=0.15,
        every axis x label/.style={at={(current axis.right of origin)},anchor=north west},
        every axis y label/.style={at={(current axis.above origin)},anchor=north east}
    }
}

\begin{tikzpicture}
\begin{axis}[
	clip bounding box=upper bound,
	legend pos=south east,
	label style={font=\footnotesize},
	tick style={font=\tiny},
	tick label style={font=\tiny},
	xlabel=$\arithaverage{\recall}$, 
	xlabel style={at={(0.96,0.17)}},
	ylabel=$\arithaverage{\precision}$, 
	ylabel style={at={(0.25,0.95)},rotate=-90},
	width=0.55\columnwidth,
	height=0.55\columnwidth,
	enlargelimits=false,
	xmin=0.5, 
	xmax=0.85, 
	ymin=0.5, 
	ymax=0.88,
	grid=major,
	minor x tick num=1,
	minor y tick num=1,
	scatter/classes={
		a={mark=x,color=black},
		b={mark=+,color=blue},
		c={mark=o,color=black!80!red},
		d={mark=asterisk,color=blue},
		e={mark=star},
		f={mark=triangle,color=orange},
		g={mark=square},
		h={mark=triangle,color=blue},
		i={mark=diamond,color=black!80!red},
		j={mark=Mercedes star,color=blue},
		k={mark=Mercedes star flipped,color=orange},
		l={mark=pentagon,color=blue},
		m={mark=oplus},
		n={mark=otimes,color=blue},
		o={mark=*},
		p={mark=square*,color=black!60!green},
		q={mark=triangle*},
		r={mark=diamond*,color=blue},
		s={mark=x,color=green},
		t={mark=x,color=red},
		u={mark=+,color=green},
		v={mark=+,color=red},
		w={mark=asterisk,color=green},
		x={mark=asterisk,color=red},
		y={mark=star,color=green},
		z={mark=square,color=red},
		ab={mark=square,color=black!40!green},
		ac={mark=triangle,color=red},
		ad={mark=triangle,color=black!80!red},
		ae={mark=diamond,color=red},
		af={mark=diamond,color=green},
		ag={mark=pentagon,color=red},
		ah={mark=pentagon,color=green},
		ai={mark=oplus,color=red},
		aj={mark=oplus,color=green},
		sbs={mark=pentagon,color=red},
		wise={mark=pentagon}
		}
	]
    \addplot[ 	scatter,only marks,mark options={scale=0.6},
		scatter src=explicit symbolic]
	table[x=R,y=P,meta=label] {cdnet-2014.dat};
\end{axis}
\end{tikzpicture}
\par\end{centering}
}\subfloat[Our procedure (PR space).\label{fig-sub:PR-ours}]{\begin{centering}
\pgfplotsset{
    standard/.style={
        axis x line=middle,
        axis y line=middle,
        enlarge x limits=0.15,
        enlarge y limits=0.15,
        every axis x label/.style={at={(current axis.right of origin)},anchor=north west},
        every axis y label/.style={at={(current axis.above origin)},anchor=north east}
    }
}

\begin{tikzpicture}
\begin{axis}[
	clip bounding box=upper bound,
	legend pos=south east,
	label style={font=\footnotesize},
	tick style={font=\tiny},
	tick label style={font=\tiny},
	xlabel=$\summarization{\recall}$, 
	xlabel style={at={(0.96,0.17)}},
	ylabel=$\summarization{\precision}$, 
	ylabel style={at={(0.25,0.95)},rotate=-90},
	width=0.55\columnwidth,
	height=0.55\columnwidth,
	enlargelimits=false,
	xmin=0.5, 
	xmax=0.85, 
	ymin=0.5, 
	ymax=0.89,
	grid=major,
	minor x tick num=1,
	minor y tick num=1,
	scatter/classes={
		a={mark=x,color=black},
		b={mark=+,color=blue},
		c={mark=o,color=black!80!red},
		d={mark=asterisk,color=blue},
		e={mark=star},
		f={mark=triangle,color=orange},
		g={mark=square},
		h={mark=triangle,color=blue},
		i={mark=diamond,color=black!80!red},
		j={mark=Mercedes star,color=blue},
		k={mark=Mercedes star flipped,color=orange},
		l={mark=pentagon,color=blue},
		m={mark=oplus},
		n={mark=otimes,color=blue},
		o={mark=*},
		p={mark=square*,color=black!60!green},
		q={mark=triangle*},
		r={mark=diamond*,color=blue},
		s={mark=x,color=green},
		t={mark=x,color=red},
		u={mark=+,color=green},
		v={mark=+,color=red},
		w={mark=asterisk,color=green},
		x={mark=asterisk,color=red},
		y={mark=star,color=green},
		z={mark=square,color=red},
		ab={mark=square,color=black!40!green},
		ac={mark=triangle,color=red},
		ad={mark=triangle,color=black!80!red},
		ae={mark=diamond,color=red},
		af={mark=diamond,color=green},
		ag={mark=pentagon,color=red},
		ah={mark=pentagon,color=green},
		ai={mark=oplus,color=red},
		aj={mark=oplus,color=green},
		sbs={mark=pentagon,color=red},
		wise={mark=pentagon}
		}
	]
    \addplot[ 	scatter,only marks,mark options={scale=0.6},
		scatter src=explicit symbolic]
	table[x=TPRC,y=PC,meta=label] {cdnet-2014.dat};
\end{axis}
\end{tikzpicture}
\par\end{centering}
}
\par\end{centering}
\caption{Summarized performances according to two different procedures, in
the cropped ROC (upper row) or PR (lower row) spaces, for 36 classifiers
evaluated on the \changedetection 2014 dataset. The performances
obtained by these procedures differ significantly. Only our summarization
procedure preserves the bijection between ROC and PR (see text).\label{fig:in-practice}}
\end{figure}
We illustrate the summarization with the \changedetection 2014 dataset~\cite{Goyette2014ANovel}
(organized in 11 categories, containing each 4 to 6 videos). We recomputed
all the performance indicators for the 36 unsupervised background
subtraction algorithms (we could consider all the algorithms as well)
whose segmentation maps are given on \url{http://changedetection.net},
compared to the ground-truth segmentation maps.

Our first experiment aims at determining if the summarized performance
is significantly affected by the summarization procedure. Figure~\ref{fig:in-practice}
compares the results of two summarization procedures, in both the
ROC and PR spaces.
\begin{enumerate}
\item In the original \changedetection procedure, performance indicators
are obtained for each category by averaging the indicators measured
on each video individually. A final summarized performance is then
computed by averaging the performance indicators over the categories.
\item We applied our summarization, with the same weights for each category
as in the original setup; thus, for each video, we have $\proba{\selectionRandomVariable=\selectionIndex}=\frac{1}{11}\times\frac{1}{M}$,
where $M$ is the number of videos in the corresponding category.
\end{enumerate}
It can be seen that the results differ significantly (both in ROC
and PR), which emphasizes the influence of the summarization procedure
for the comparison between algorithms. However, with the original
summarization procedure, the intrinsic relationships between indicators
are not preserved at the summarized level, while ours preserves them.

Our second experiment aims at determining if the ranking between algorithms
depends on the summarization procedure. Because the F-score is often
considered as an appropriate ranking indicator for background subtraction,
we performed our experiment with it. Table~\ref{tab:ranking-discrepancies}
shows the $\Fscore$-scores and ranks obtained according to the same
two summarization procedures. We see that the summarization procedure
affects the ranking.
\begin{table}
\noindent \begin{centering}
{\footnotesize{}}%
\begin{tabular}{|c|c|c|}
\hline 
{\footnotesize{}Algorithm} & {\footnotesize{}$\Fscore$ of \cite{Goyette2014ANovel}} & {\footnotesize{}$\Fscore$ {[}ours{]}}\tabularnewline
\hline 
\hline 
{\footnotesize{}SemanticBGS} & \textbf{\footnotesize{}0.8098 (1)} & \textbf{\footnotesize{}0.8479 (1)}\tabularnewline
\hline 
{\footnotesize{}IUTIS-5} & \textcolor{red}{\footnotesize{}0.7821 (2)} & \textcolor{blue}{\footnotesize{}0.8312 (3)}\tabularnewline
\hline 
{\footnotesize{}IUTIS-3} & \textcolor{blue}{\footnotesize{}0.7694 (3)} & \textcolor{magenta}{\footnotesize{}0.8182 (5)}\tabularnewline
\hline 
{\footnotesize{}WisenetMD} & \textcolor{teal}{\footnotesize{}0.7559 (4)} & {\footnotesize{}0.7791 (10)}\tabularnewline
\hline 
{\footnotesize{}SharedModel} & \textcolor{magenta}{\footnotesize{}0.7569 (5)} & {\footnotesize{}0.7885 (8)}\tabularnewline
\hline 
{\footnotesize{}WeSamBE} & {\footnotesize{}0.7491 (6)} & {\footnotesize{}0.7792 (9)}\tabularnewline
\hline 
{\footnotesize{}SuBSENSE} & {\footnotesize{}0.7453 (7)} & {\footnotesize{}0.7657 (12)}\tabularnewline
\hline 
{\footnotesize{}PAWCS} & {\footnotesize{}0.7478 (8)} & \textcolor{teal}{\footnotesize{}0.8272 (4)}\tabularnewline
\hline 
\end{tabular}
\par\end{centering}
\caption{Extract of $\protect\Fscore$-scores (ranks) obtained with two summarization
procedures on \changedetection.\label{tab:ranking-discrepancies}}
\end{table}

\textcolor{red}{}

\section{Conclusion}

\label{sec:conclusion}

In background subtraction, algorithms are evaluated by applying them
on videos and comparing their results to ground-truth references.
Performance indicators of individual videos are then combined to derive
indicators representative for a set of videos, during a procedure
called ``summarization''.

We have shown that a summarization procedure based on the arithmetic
mean leads to inconsistencies between summarized indicators and complicates
the comparison between algorithms. Therefore, based on the definition
of a random experiment, we presented a new summarization procedure
and formulas that preserve the intrinsic relationships between indicators.
These summarized indicators are specific for this random experiment
and for the weights given to the various videos. An easy-to-code algorithm
for our summarization procedure has been given. Our procedure is illustrated
and commented on the \changedetection dataset.

As a general conclusion, for background subtraction involving multiple
videos, we recommend to always use our summarization procedure, instead
of the arithmetic mean, to combine performance indicators calculated
separately on each video. By doing so, the formulas that hold between
them at the video level also hold at the summarized level.

\clearpage{}

\end{document}